%% file: main.tex
\documentclass[10pt,twocolumn,letterpaper]{article}

\usepackage{iccv}              

\newcommand{\methodname}{LGDiST}
\newcommand{\specialgenes}{HSAGs}


\definecolor{iccvblue}{rgb}{0.21,0.49,0.74}
\usepackage[pagebackref,breaklinks,colorlinks,allcolors=iccvblue]{hyperref}
\usepackage{xcolor}
\usepackage{float}
\usepackage{marvosym}

\def\paperID{44} 
\def\confName{ICCV}
\def\confYear{2025}

\title{Latent Gene Diffusion for Spatial Transcriptomics Completion}

\author{Paula Cárdenas\textsuperscript{\Letter}, Leonardo Manrique$^*$, Daniela Vega$^*$, Daniela Ruiz, Pablo Arbeláez\\
Center for Research and Formation in Artificial Intelligence\\
Universidad de los Andes, Bogotá, Colombia\\
{\tt\small \{p.cardenasg,dl.manrique,d.vegaa,da.ruizl1,pa.arbelaez\}@uniandes.edu.co}
}

\begin{document}
\maketitle
\def\thefootnote{*}\footnotetext{These authors contributed equally to this work}\def\thefootnote{\arabic{footnote}}
\input{sec/0_abstract}   
\input{sec/1_intro}
\input{sec/2_related_work}

\input{sec/3_LGDiST}

\input{sec/4_results}
\input{sec/5_conclusions}
\input{sec/6_acknowledgments}
{
    \small
    \bibliographystyle{ieeenat_fullname}
    \bibliography{main}
}

\end{document}

%% file: sec/0_abstract.tex
\begin{abstract}
Computer Vision has proven to be a powerful tool for analyzing Spatial Transcriptomics (ST) data. However, current models that predict spatially resolved gene expression from histopathology images suffer from significant limitations due to data dropout. Most existing approaches rely on single-cell RNA sequencing references, making them dependent on alignment quality and external datasets while also risking batch effects and inherited dropout. In this paper, we address these limitations by introducing \methodname{}, the first reference-free latent gene diffusion model for ST data dropout. We show that \methodname{} outperforms the previous state-of-the-art in gene expression completion, with an average Mean Squared Error that is 18\% lower across 26 datasets. Furthermore, we demonstrate that completing ST data with \methodname{} improves gene expression prediction performance on six state-of-the-art methods up to 10\% in MSE. A key innovation of \methodname{} is using context genes previously considered uninformative to build a rich and biologically meaningful genetic latent space. Our experiments show that removing key components of \methodname{}, such as the context genes, the ST latent space, and the neighbor conditioning, leads to considerable drops in performance. These findings underscore that the full architecture of \methodname{} achieves substantially better performance than any of its isolated components.
\end{abstract}

%% file: sec/1_intro.tex
\section{Introduction}
\label{sec:intro}

The emergence of Spatial Transcriptomics (ST) is one of the most groundbreaking advances in biological research over the last five years \cite{marx2021method}. This technology bridges the gap between genetic expression and cellular spatial organization by combining single-cell RNA sequencing (scRNA-seq) and histology in a single tool \cite{wang2023spatial}. ST technologies can be broadly categorized into imaging-based and sequencing-based methods. Imaging-based technologies achieve single-cell resolution but have very limited sequence coverage \cite{wang2023spatial}, whereas sequencing-based technologies have a low spot-level resolution but offer high throughput \cite{cheng2023spatially}. In this work, we focus on sequencing-based ST technologies, which measure gene expression in spatially localized bulk regions rather than at the level of individual cells.

Despite its advantages, ST is a resource-demanding technology that involves expensive devices, complex sample handling, and is time-consuming \cite{jin2024advances}. In response, ongoing research has introduced multiple Computer Vision (CV) models to automate ST by predicting a tissue's spatially resolved genetic expressions solely from histopathology images \cite{bleep}. Proposed gene expression prediction models leverage a wide range of CV techniques and Artificial Intelligence (AI) algorithms with varied complexities, from simple Convolutional Neural Networks (CNNs) \cite{stnet} to Graph Neural Networks (GNNs) \cite{yang2024spatial} and contrastive learning techniques \cite{egn,bleep}. Nevertheless, another critical but frequently overlooked limitation of ST is data dropout. This phenomenon refers to the stochastic failure to detect gene transcripts in a spot, even when they are expressed \cite{dropout}. Dropouts introduce noise into ST data and hinder the utility of this technology. Moreover, noise in ST datasets not only affects genetic analyses but seriously degrades the performance of CV models trained for a wide variety of image-processing tasks, including gene expression prediction from histopathology images, spatial domain detection, cellular segmentation and classification, and computer-aided diagnosis \cite{li2022emerging,defard2024point,komura2024machine}.

\begin{figure*}[t]
\includegraphics[width=\textwidth]{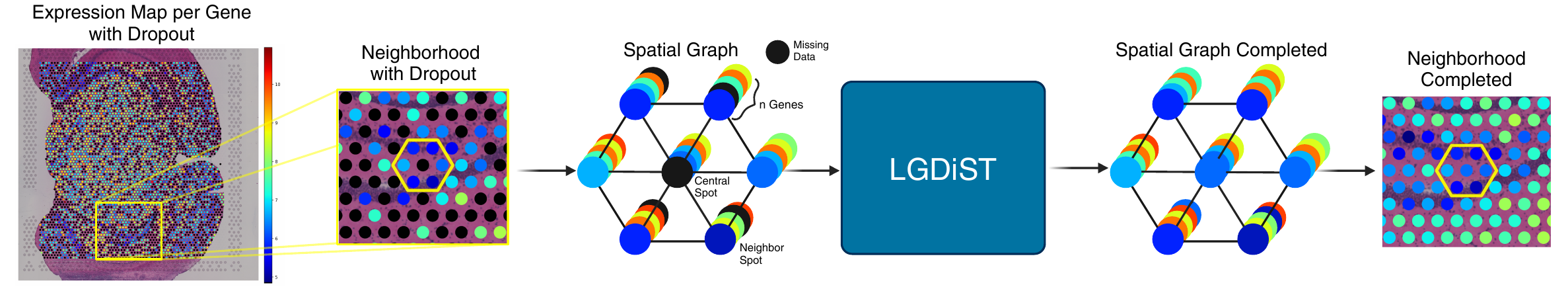}
\caption{\textbf{Spatial Transcriptomics Completion Pipeline.} \methodname{} processes ST neighborhoods to infer missing values, generating a completed expression map with high-quality data.} 
\label{fig:visual_abstract}
\end{figure*}

The dropout problem has received increasing attention in recent years due to its significant impact on the effective processing of ST data \cite{stPlus,stDiff,spage, tangram,scDiffusion}. However, most of these works are primarily suitable for imaging-based technologies, given their inherent focus on single-cell information. This paradigm limits the applicability of these methods for data completion and leaves the dropout problem unresolved for ST data measured through bulk transcriptomics. Additionally, most methods previously proposed are based on scRNA-seq references to accurately predict missing genetic values. This reliance not only makes the models highly sensitive to the quality of external data but also increases the challenge of addressing dropout in sequencing-based ST, given its lower resolution.

To overcome the limitations mentioned above, this paper introduces Latent Gene Diffusion for Spatial Transcriptomics (\methodname{}), a novel reference-free generative model designed to complete missing gene expression values in ST data. Our approach builds on the remarkable success of Denoising Diffusion Probabilistic Models (DDPMs) \cite{ddpms} in image generation and biology-related applications, including small molecule generation, protein structure prediction \cite{soleymani2024structure}, and, more recently, gene expression prediction from histopathology images \cite{stem}. To enhance our method, we also draw inspiration from the ability of AutoEncoders (AEs) to capture complex patterns in data and build highly informative latent spaces \cite{berahmand2024autoencoders}. \methodname{} utilizes an AE to build an ST latent space incorporating Highly Spatially Associated Genes (HSAGs) and additional genes with lower Moran's I, which we consider as context genes (CGs). Furthermore, we employ a Diffusion Transformer (DiT) \cite{dit} conditioned on the encoded genetic data from the neighboring spots to complete missing values in the central spot. 

We perform multiple comparative experiments across the 26 sequencing-based ST datasets in the SpaRED database \cite{spackle}, assessing our model against leading methods on dropout data completion. The results show a noteworthy improvement in the prediction of missing data with an average 18\% reduction in Mean Square Error (MSE) across all the datasets explored compared to the state-of-the-art in reference-free ST data completion. Additionally, we examine the significance of \methodname{} by quantifying its impact on the performance of models designed to predict total gene expression profiles from histopathology images. We find that \methodname{} effectively cleans up training data, allowing these state-of-the-art models to learn from higher quality information and subsequently improve their own evaluation metrics up to 10\%. These results establish \methodname{} as a superior approach for addressing data dropout in ST and highlight its potential to enhance downstream CV analyses, ultimately advancing the accuracy and reliability of spatial transcriptomics studies.

To summarize, our main contributions are: 
\begin{enumerate}
    \item We introduce the first generative, reference-free model for addressing data dropout in ST and prove its ability to optimize the performance of CV models trained on ST data. 
    \item We design a latent space for ST data and show that conducting diffusion within it enhances gene expression recovery. 
    \item We demonstrate the value of low spatially autocorrelated genes in spatial analyses, presenting a new way of harnessing data previously considered uninformative. 
\end{enumerate}
To promote further research on ST, our source code is publicly available at \url{https://github.com/BCV-Uniandes/LGDiST}. 

%% file: sec/2_related_work.tex
\section{Related Work}
\label{sec:related_work}
To mitigate the effect of data dropout, previous works have proposed various methods that aim to complete missing values in ST datasets. These methods can generally be divided into two main categories based on whether they require external references or rely solely on the ST data itself: reference-based and reference-free approaches. Each category has distinct advantages and limitations, which have driven the development of increasingly sophisticated strategies to address the inherent challenges posed by dropout in spatial transcriptomics.

\begin{figure*}[t]
\includegraphics[width=\textwidth]{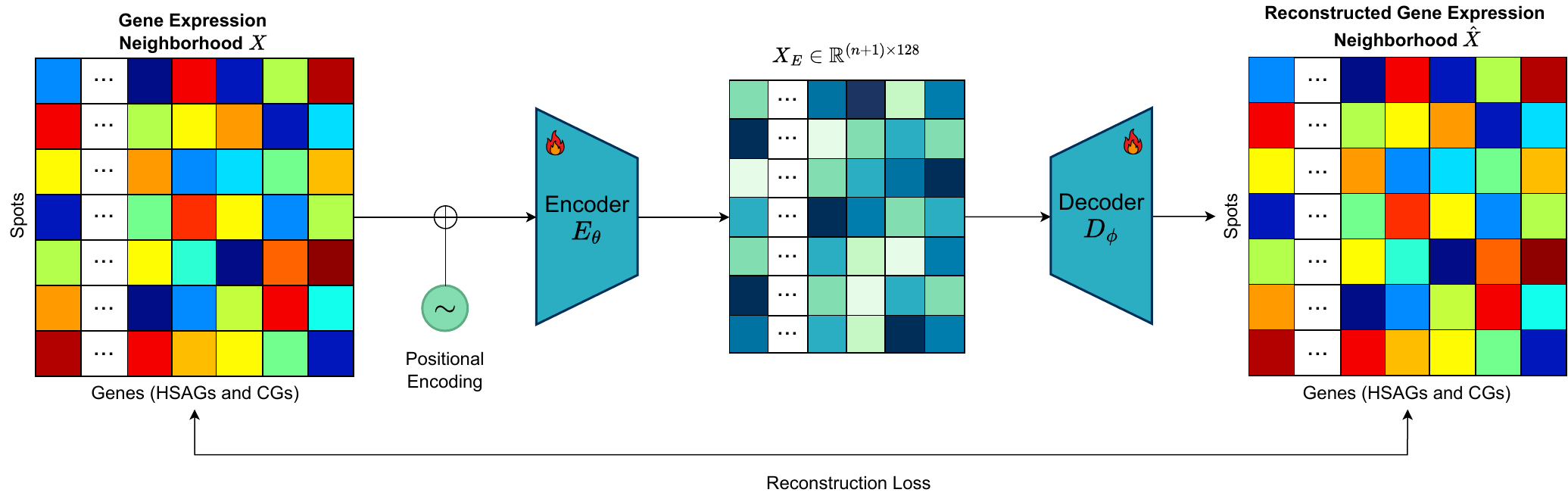}
\caption{\textbf{AutoEncoder training framework}. We add a 2D positional encoding to the gene expression neighborhood $X$ before passing it through an encoder $E_{\theta}$ that maps it to a Latent Space Representation $X_{E}$. The decoder $D_{\phi}$ then reconstructs the input from the latent space, aiming to minimize reconstruction loss.} 
\label{fig:autoencoder}
\end{figure*}

\subsection{Reference-based methods}
Reference-based methods \cite{spage,stPlus,stDiff} attempt to enhance the completeness of ST datasets by leveraging information from scRNA-seq data as an external reference. By aligning gene expression profiles measured in spatial spots with those obtained from single-cell technologies, these methods aim to predict the missing values on the expression map of each gene. A representative example of this strategy is stDiff \cite{stDiff}, which utilizes DDPMs \cite{ddpms} to model the relationships between gene expression abundances based on scRNA-seq-derived references. This modeling enables highly informed predictions of missing expression values based on prior knowledge of genetic expression profiles. 

Nonetheless, incorporating scRNA-seq references into the completion process significantly increases the computational resources and experimental data requirements, since high-quality scRNA-seq datasets must be collected, processed, and aligned with the ST samples. Moreover, differences in sequencing technologies or protocols can introduce inconsistencies or batch effects that propagate into the completed ST data, compromising its reliability and introducing biases that may affect downstream analyses \cite{marel2024navigating}. Importantly, reference-based models depend on alignment quality, which remains suboptimal despite ongoing advancements, potentially introducing bias in the completed data \cite{yan2024integration}. Recognizing these limitations, we propose a model that works entirely without scRNA-seq references, thereby eliminating dependence on external datasets and avoiding risks associated with alignment performance. This approach makes \methodname{} broadly applicable to bulk ST data and suitable for tissues lacking existing reference information, while minimizing potential biases.

\subsection{Reference-free methods} \label{sec:reference-free}
Reference-free methods \cite{stLearn} complete missing values using only the information contained within the ST dataset itself. These methods avoid dependence on external references and are often more adaptable to a wider range of samples and experimental conditions. 
In \cite{sepal}, authors propose a median-completion strategy that operates as a reference-free method. This method uses a modified adaptive median filter to replace dropout values with the median expression of the gene of interest in a local region surrounding the missing spot. In case the region lacks sufficient data, the missing values are substituted with the median value of the whole slide image (WSI). This strategy proved to be helpful as an initial approach to reduce the effect of data dropout for gene prediction methods. Despite its practicality, relying on median imputation alone can excessively smooth the expression patterns of genes across the tissue, diminishing biologically relevant spatial variability. Furthermore, median completion lacks the robustness required to handle datasets with a high fraction of dropout, where too many missing values make the median estimate unreliable. 

More advanced reference-free methods, such as SpaCKLE \cite{spackle}, address the dropout problem by adopting a masked autoencoder architecture that learns patterns of gene expression directly from the spatial data. SpaCKLE represents a significant advancement, utilizing a deep learning framework and achieving superior performance compared to previous methods. However, it focuses exclusively on HSAGs, which are selected based on their Moran’s I scores \cite{moran}. This narrow focus overlooks genes with low spatial autocorrelation, which can still provide valuable context and biological insights. By discarding these low-autocorrelation genes, SpaCKLE limits its capacity to capture a broader and more comprehensive molecular context within the tissue. 

In contrast, \methodname{} explicitly incorporates a latent space that encompasses both HSAGs and low-autocorrelation genes that serve as context. This design choice enables our model to leverage a richer set of genetic information, improving its ability to reconstruct missing expression values in a way that better preserves both local and global spatial expression patterns.

%% file: sec/3_LGDiST.tex
\section{Latent Gene Diffusion}
\label{sec:LGDiST}

\subsection{Overview}

\begin{figure*}[t]
\includegraphics[width=\textwidth]{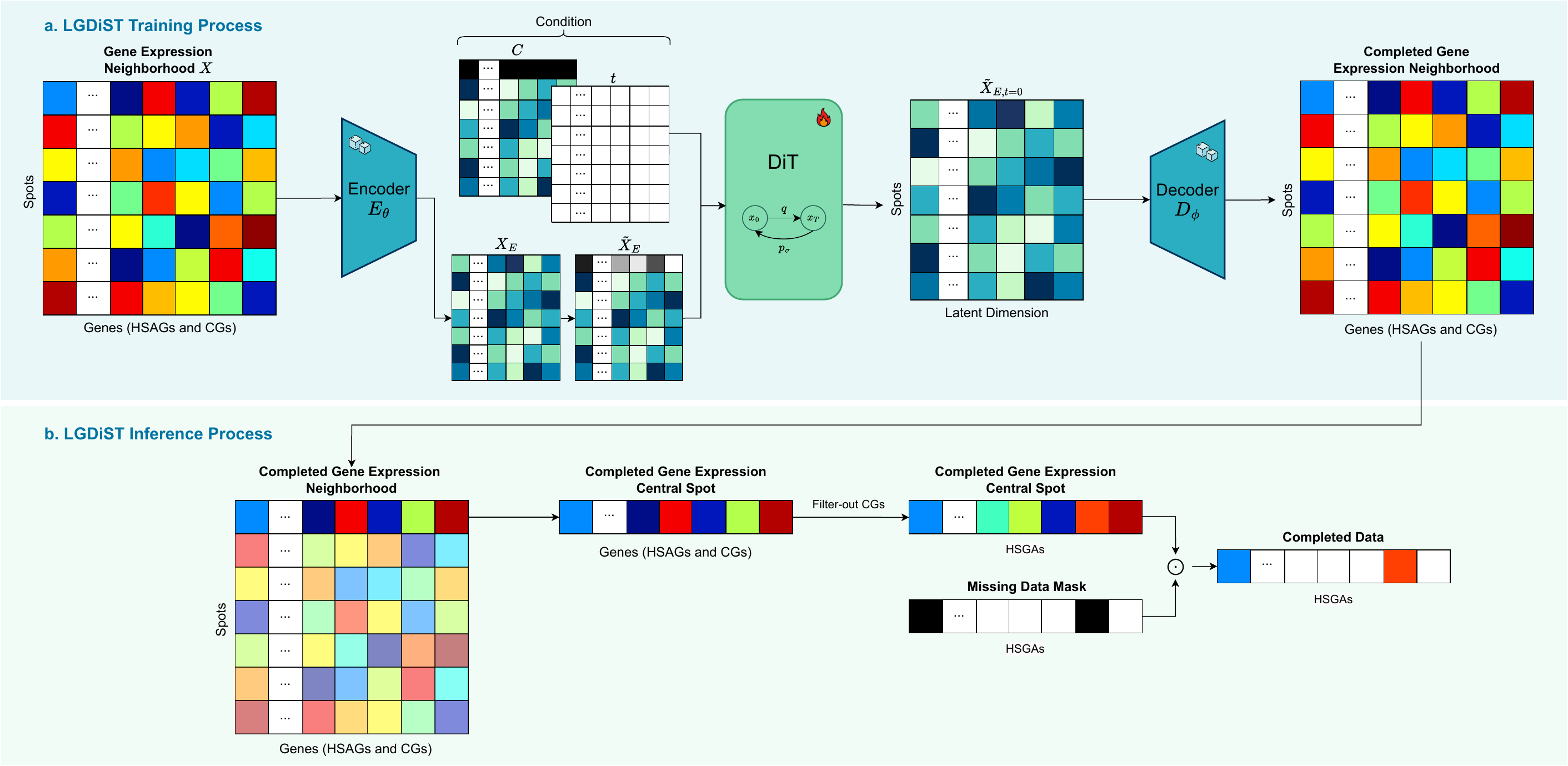}
\caption{(a) \methodname{} Training: The encoder \(E_{\theta}\) maps the gene expression neighborhood $X$ to $X_E$. We feed this representation to the DiT with the condition $C$ and $t$. Through an iterative noising-denoising process, the DiT learns to predict the noise in $\tilde{X}_{E, t}$. After learning how to remove the noise, the DiT outputs $\tilde{X}{E, t=0}$ and $D_{\phi}$ decodes the result to obtain the final gene expression matrix. (b) \methodname{} Inference: After obtaining the completed gene expression neighborhood, we filter out the CGs and obtain the final completed data.} 
\label{fig:lgdist}
\end{figure*}

\methodname{} consists of two main components: an AE (shown in Fig. \ref{fig:autoencoder}) that builds a compact but highly informative latent space for ST data, and a diffusion model (Fig. \ref{fig:lgdist}) trained to generate the missing gene expression profile of a spot conditioned on the information from its nearest neighbors. 

We define $X \in \mathbb{R}^{(n+1) \times g}$ as the matrix representing the genetic expression profile of a neighborhood, where $g$ is the number of genes and $n$ denotes the number of closest neighbors for a given spot in the hexagonal geometry of Visium ST data. Consequently, we construct a neighborhood for each spot to consider its local context during data processing.

Our methodology follows a structured two-stage training process. First, we train the AE to build a latent space for ST neighborhoods. Once trained, we use this model to encode and decode the input and output of the DiT, respectively. During the second stage, we train the DiT conditioned on the latent representation of ST neighborhoods. This approach follows a latent diffusion strategy as proposed in \cite{LDM}, which accelerates training and reduces memory requirements during both training and inference in the diffusion framework. Thus, instead of employing existing genetic foundation models to obtain gene expression embeddings, which may not capture the neighborhood-level structure our framework requires, we design and train our own AE that both embeds the expression profiles and decodes them once the diffusion process is complete.

\subsection{Genomics-Based Latent Space}

We propose an autoencoder (AE) architecture consisting of a four-layer transformer-based encoder with a single attention head, followed by a two-layer multilayer perceptron (MLP) decoder. Both the encoder and decoder use a dropout rate of 0.1 and the hyperbolic tangent activation function. As shown in Fig. \ref{fig:autoencoder}, we feed our encoder with $X_{p}$, which corresponds to the gene expression matrix $X$ with added 2D positional encodings. Then we map $X_{p}$ to the latent space, obtaining \( X_{E} \in \mathbb{R}^{(n+1) \times d} =  E_{\theta}(X_{p})\), where $d$ corresponds to a latent dimension of $128$. To enhance robustness, we apply a regularization mechanism that injects noise into $X_{p}$ during training with a probability of 50\%. For decoding, we use the MLP \( D_{\phi} \) to reconstruct the expression profile of the processed neighborhood based on its latent representation \( X_{E} \). The final reconstructed expression matrix corresponds to \( \hat{X} \in \mathbb{R}^{(n+1) \times g} = D_{\phi}(X_{E})\).

We train the autoencoder to minimize an adjusted reconstruction loss \( \mathcal{L}_{\text{rec}} \) that prioritizes the columns corresponding to the \specialgenes~through the use of a weighted factor $\alpha$:

\begin{equation}
    \mathcal{L}_{\text{HSAGs}} = MSE(X_{\text{HSAGs}}, \hat{X}_{\text{HSAGs}})
\end{equation}
\begin{equation}
    \mathcal{L}_{\text{CGs}} = MSE(X_{\text{CGs}}, \hat{X}_{\text{CGs}})
\end{equation}
\begin{equation}
    \mathcal{L}_{\text{rec}} = \alpha \cdot \mathcal{L}_{\text{HSAGs}} + (1-\alpha) \mathcal{L}_{\text{CGs}}
\end{equation}

\subsection{Completing Missing Data with DDPMs}

During the second stage of our model, we train a twelve-layer DiT with sixteen attention heads to predict the expression values of all genes in the central spot using information from its nearest neighbors. We define the output of each noising step in our diffusion process as described in Eq. \ref{eq:noised_input}. Here, $\varepsilon \sim \mathcal{N}(0, I)$ is Gaussian noise, and $M \in \{0,1\}^{(n+1) \times d}$ is a binary mask that ensures only the central spot’s row is noised while leaving the neighboring spots' data unchanged. The model learns to progressively denoise $\tilde{X}_{E}$ by estimating the added noise and removing it iteratively. As shown in Fig. \ref{fig:lgdist}a, we condition the diffusion process on $C= M \odot X_{E}$ to ensure that the model relies solely on the neighborhood information when reconstructing the central spot’s gene expression profile.

\begin{equation}
    \tilde{X}_{E} = (1 - M) \odot \varepsilon + M \odot X_{E}
    \label{eq:noised_input}
\end{equation}

The model is trained to minimize the standard diffusion loss in Eq. \ref{eq:diffusion_loss_eq}, where $\hat{\varepsilon}_\theta(\tilde{X}_{E, t}, C, t)$ is the model’s predicted noise at timestep $t$ based on the input noise $\tilde{X}_{E, t}$ and the condition $C$.

\begin{equation}
    \mathcal{L}_{DDPM} = \mathbb{E}_{X_{E}, t, \varepsilon} \left[ \| \varepsilon - \hat{\varepsilon}_\theta(\tilde{X}_{E, t}, C, t) \|_{2}^2 \right] 
    \label{eq:diffusion_loss_eq}
\end{equation} 

After training our model to complete fully empty central spots, we use it to tackle the real dropout problem, where only a subset of genes in the central spot's profile is missing. As shown in Fig. \ref{fig:lgdist}b, during inference we must decode the DiT's generated expressions and filter out CGs. Notably, instead of retrieving the whole generated profile, we extract only the values corresponding to the missing data. On the other hand, when evaluating our model, we build a mask that simulates data dropout by hiding values that were successfully measured by the ST technology. This procedure allows us to directly compare the predicted values against the known ground truth measurements and compute reliable evaluation metrics.

\subsection{Implementation Details}

Given the substantial amount of missing values in the original ST data, we adopt the training strategy outlined in \cite{spackle}. This strategy includes pre-completing the data before training using the median completion technique from \cite{sepal}, as detailed in section \ref{sec:reference-free}. We adopt this approach since \cite{spackle} demonstrates that data pre-completion accelerates training convergence and ensures meaningful, non-zero outputs. Furthermore, \cite{spackle} shows that pre-completion improves the performance of the final gene completion model compared to training on the original data with missing values. This strategy also aligns with the reference-free nature of our approach, as the median-completion technique fills missing values using information already contained in the ST dataset.

We train our model on an NVIDIA Quadro RTX 8000, using an AdamW optimizer \cite{kingma2014adam}, and a batch size of 128. The DiT backbone requires approximately 3.22 GFLOPs, and we train it for 1500 epochs with a cosine noise scheduler and a learning rate of $1 \times 10^{-4}$, with 1500 diffusion steps during training and 50 diffusion steps for sampling. The autoencoder follows a longer training process of 5000 epochs, with a learning rate of $1 \times 10^{-6}$.

%% file: sec/4_results.tex
\section{Experimental Results}
\subsection{Experimental Setup}
We conduct experiments using the 26 publicly available datasets in SpaRED \cite{spackle}, for both data completion and gene expression prediction from histopathology images. The exact details of the dataset composition are provided on the \href{https://bcv-uniandes.github.io/spared_webpage/}{SpaRED} website. To focus only on genes with strong spatial autocorrelation, SpaRED filters ST data using Moran’s I, retaining either 32 or 128 genes with the highest spatial autocorrelation scores for each dataset. We use SpaRED's suggested genes as our set of HSAGs. To include additional context genes (CGs), we adopt the same pre-processing strategy as described in \cite{spackle}, but we adjust the filtering criteria to select 1,024 genes based on Moran's I. This process preserves each dataset's original 32 or 128 HSAGs while integrating the additional informative context genes. We evaluate performance by computing the standard MSE metric and Pearson Correlation Coefficient (PCC) between the ground truth and the completed or predicted expression values.

\begin{figure}[t]
    \centering
    \includegraphics[width=0.95\linewidth]{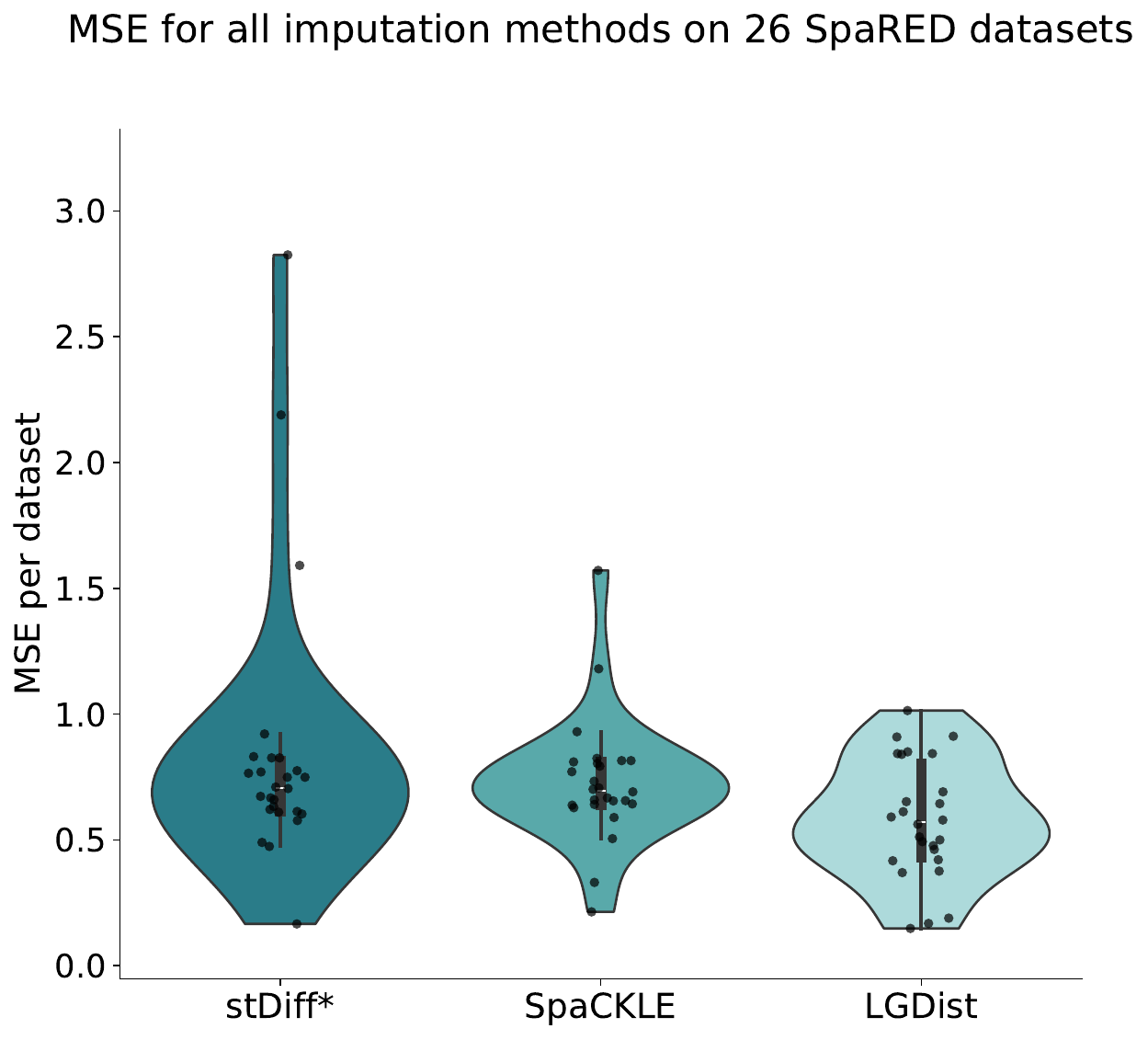}
    \caption{MSE of the three completion methods on each dataset. stDiff* corresponds to an adjusted version of stDiff \cite{stDiff}, modified to work reference-free.}
    \label{fig:imputation_quantitative_comparison}
\end{figure}


\begin{figure*}[t]
\includegraphics[width=\textwidth]{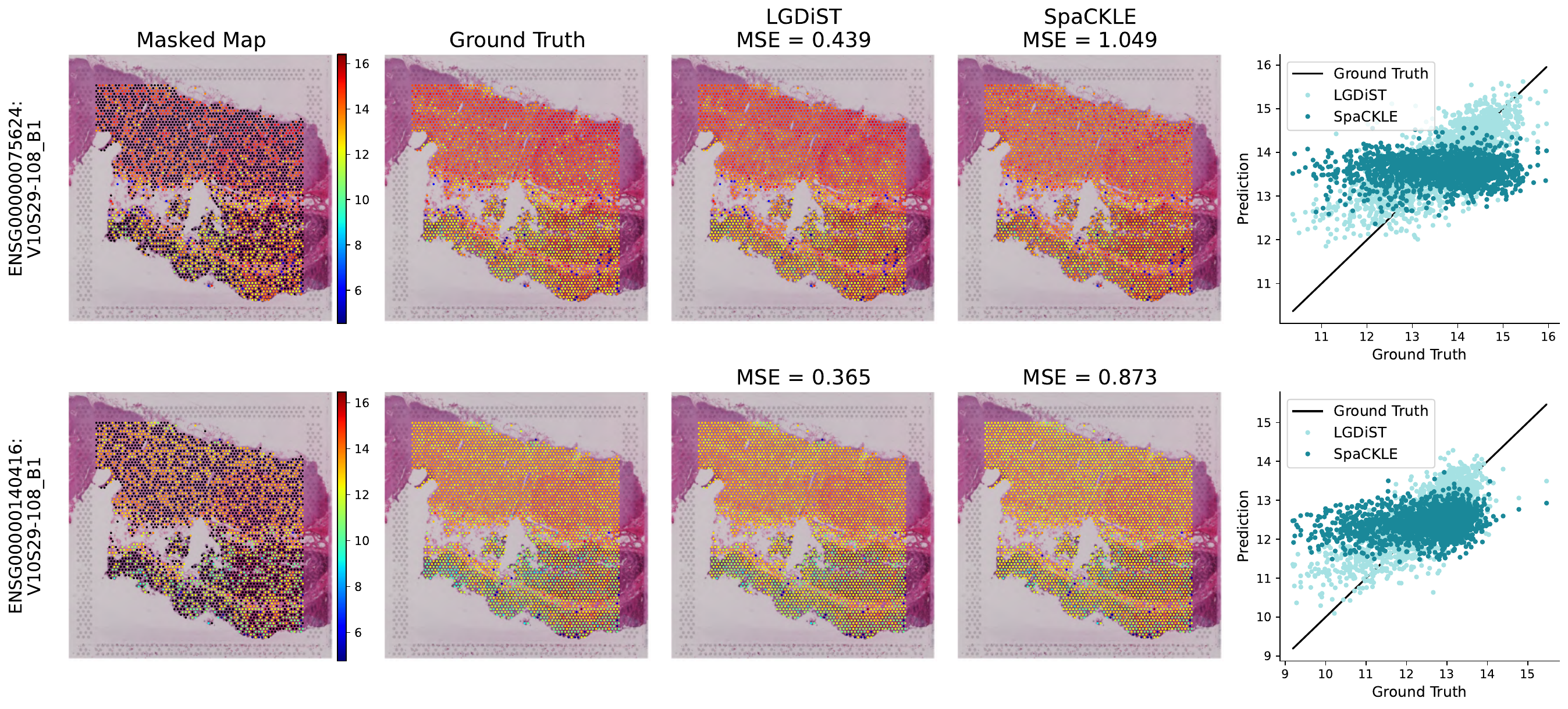}
\caption{Optimal qualitative results for data completion with \methodname{} vs with SpaCKLE \cite{spackle}.} 
\label{fig:optimal_qualitative}
\end{figure*}

\subsection{Missing Data Completion}
We compare the performance of \methodname{} on the data completion task against SpaCKLE \cite{spackle}, the previous state-of-the-art among reference-free methods, and a modified version of stDiff \cite{stDiff}, which, like our method, employs DiT for data completion. To ensure a fair comparison, we adapt stDiff to work exclusively with ST data by restructuring its diffusion conditioning, ensuring it no longer relies on external information.


Fig. \ref{fig:imputation_quantitative_comparison} displays the distribution of the MSE values obtained by the three models when completing missing data on each one of the 26 datasets. These results demonstrate that \methodname{} consistently outperforms the previous state-of-the-art with an average MSE 18\% lower than SpaCKLE's. Moreover, despite stDiff also being based on DiT, the sum of our workflow strategies proves to maximize the potential of the generative model.

\begin{figure*}[t]
\includegraphics[width=\textwidth]{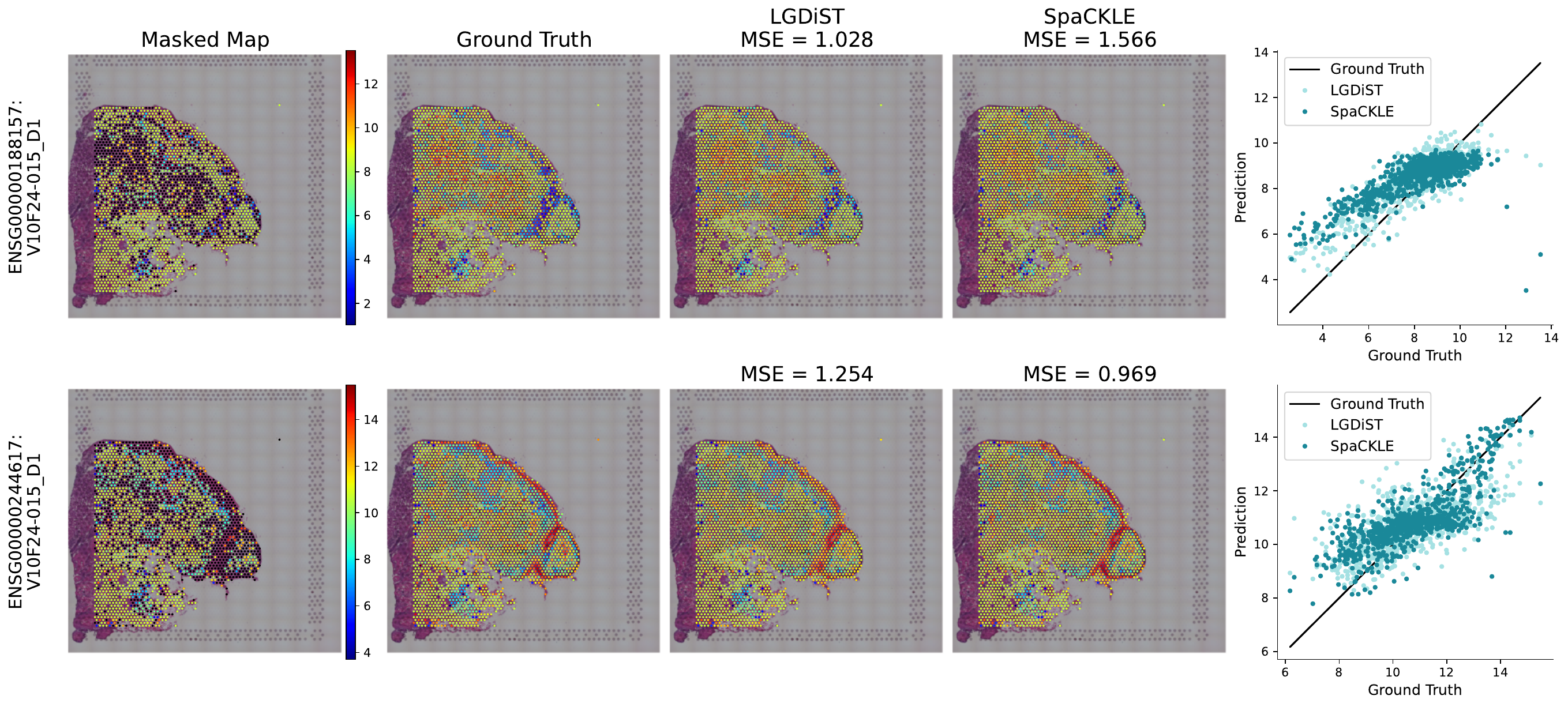}
\caption{Suboptimal qualitative results for data completion with \methodname{} vs with SpaCKLE \cite{spackle}.} 
\label{fig:suboptimal_qualitative}
\end{figure*}

In Fig. \ref{fig:optimal_qualitative}, we present the result of completing simulated missing values (black dots on the masked map) for two different genes with optimal results for \methodname{}. Our method completes missing data without degrading the general spatial pattern followed by a gene. Unlike SpaCKLE \cite{spackle}, \methodname{}'s predictions clean the noisy expression map without inducing data sparsity or smoothing. The scatter plots in Fig. \ref{fig:optimal_qualitative} further demonstrate that the data completed with \methodname{} exhibits a distribution closer to the ground-truth than the data completed with SpaCKLE \cite{spackle}. 

On the other hand, Fig. \ref{fig:suboptimal_qualitative} presents examples of suboptimal performance from our model. In these cases, \methodname{}'s predictions deviate further from the target values, and, in some cases, align more closely with the median expression in the tissue. The compression of \methodname{}'s predicted values toward a narrow expression range in the scatter plot reflects this behavior. This tendency is also noticeable in the predicted expression map of the second row, where the red spots have a lighter tone than the ground truth and present an underestimation of the gene expression in that location. However, despite reducing the range of expression in these suboptimal results, \methodname{} is still capable of preserving spatial patterns. 

\begin{figure}
    \centering
    \includegraphics[width=0.95\linewidth]{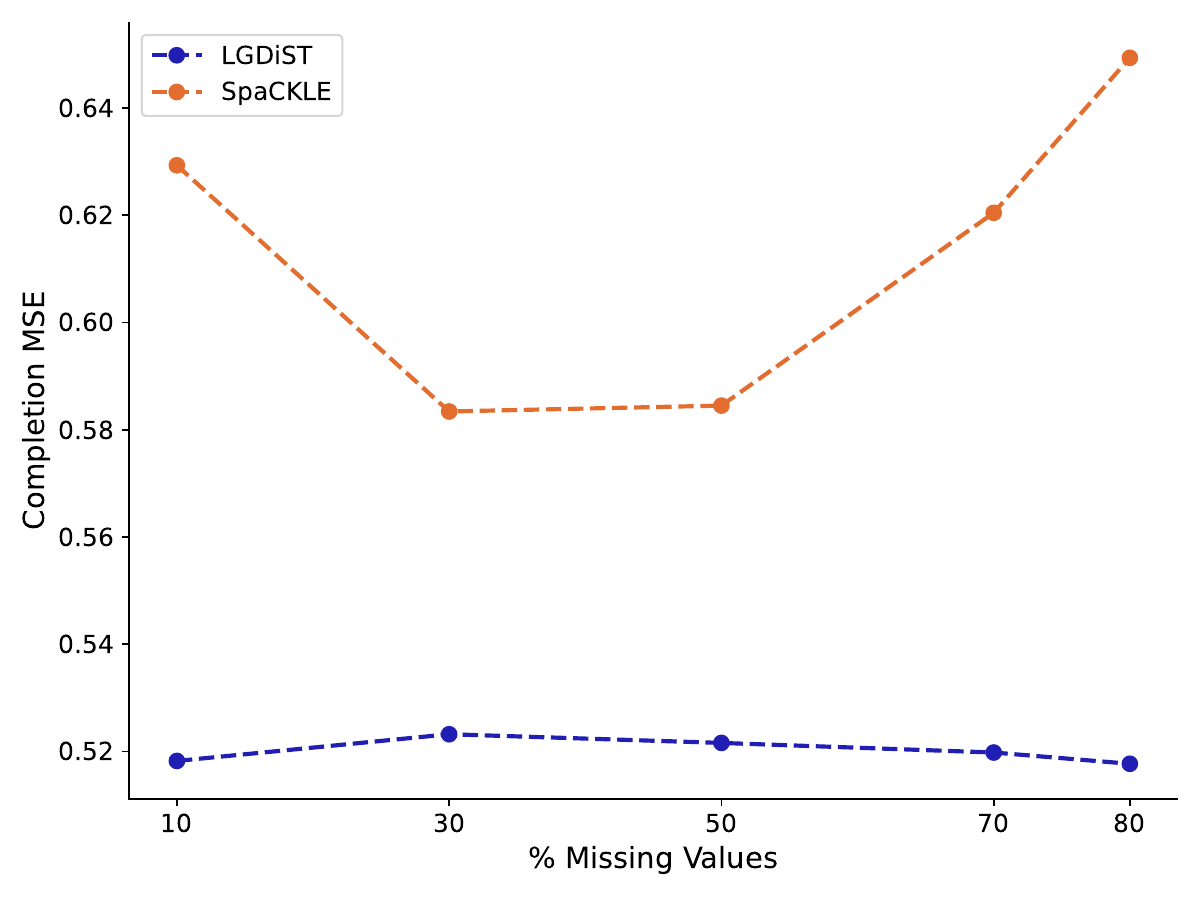}
    \caption{Line plot displaying completion MSE for \methodname{} and SpaCKLE \cite{spackle} methods across different percentages of synthetically masked data.}
    \label{fig:robustness_results}
\end{figure}

To evaluate the robustness of our model, we explore the completion results with a varied fraction of masked values that act as synthetically missing data. Fig. \ref{fig:robustness_results} illustrates the MSE as a function of the percentage of missing data for \methodname{} and SpaCKLE in the 10XGMBSP data set. Across all percentages, \methodname{} consistently achieves a lower MSE compared to SpaCKLE, demonstrating its superior performance in handling incomplete data. SpaCKLE’s error increases sharply at higher levels of missing data (70\% and 80\%), whereas \methodname{} maintains a stable and low MSE even as the task becomes more challenging.

\subsection{Gene Prediction Models Enhancement}
\begin{figure*}
\centering
\includegraphics[width=\textwidth]{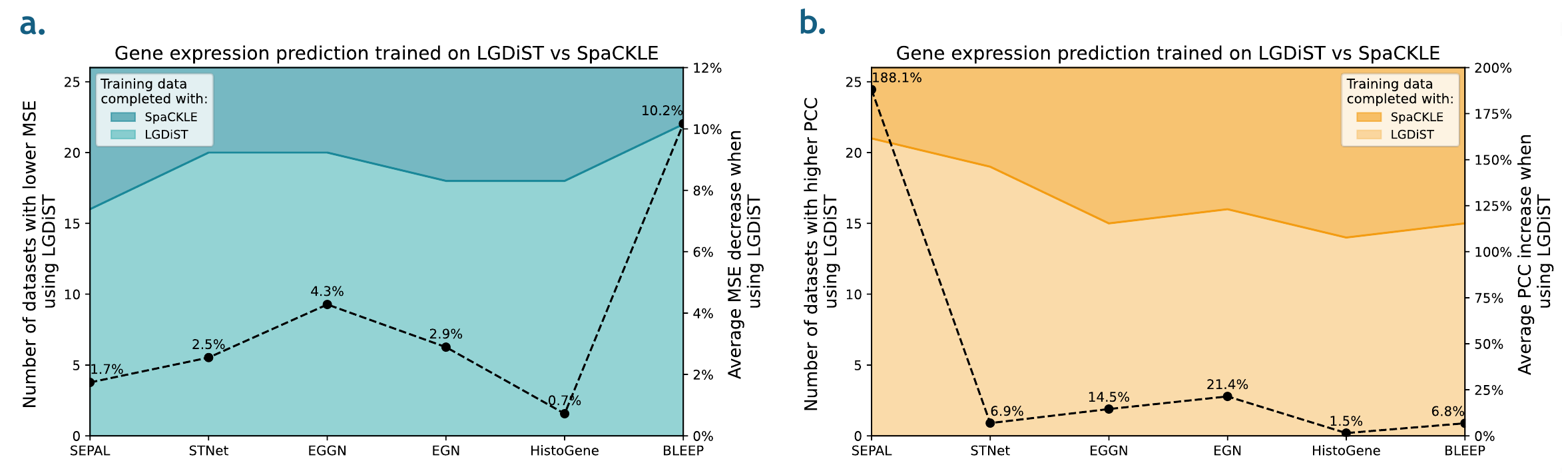}
\caption{(a) MSE and (b) PCC of gene-expression prediction models when trained on \methodname{}-completed data compared to training on data completed with SpaCKLE \cite{spackle}.}
\label{fig:significance_results}
\end{figure*}

To further evaluate the significance of our contributions, we use six models previously proposed in the literature to tackle the gene expression prediction task \cite{stnet,histogene,egn,yang2024spatial,sepal,bleep}. We assess these models' ability to predict gene expression profiles from histopathology images when trained on data whose dropout was completed by either SpaCKLE \cite{spackle} or \methodname{}. Following \cite{spackle}, during testing, we compute the evaluation metric (MSE) only on gene expression values that were directly measured by ST technology, excluding those that were completed. This ensures that performance improvements are not artificially influenced by the completion method itself, but rather reflect a genuine enhancement in the models' predictive capabilities.



Fig. \ref{fig:significance_results} compares the (a) MSE and (b) PCC, of gene expression prediction models when trained with \methodname{} and SpaCKLE \cite{spackle}. The left Y-axis displays the number of datasets in which each prediction method improved, while the right Y-axis indicates the improvement percentage in the corresponding evaluation metric when training on \methodname{}-completed data versus SpaCKLE-completed data \cite{spackle}.
These results indicate that all models achieve better gene expression prediction in more than half of the datasets when trained on \methodname{}-completed data. Additionally, all six models exhibit an average improvement in MSE and PCC, confirming the benefit of training on dropout-corrected data using \methodname{}. 

The reduction in MSE indicates that the predicted gene expression values are more accurate and closer to the true expression levels when trained on \methodname{} completed data. Notably, BLEEP demonstrates the greatest improvement, achieving an average MSE reduction of 10\%. Furthermore, the improvement in PCC reflects a stronger preservation of spatial patterns in the predicted gene expression values. This is crucial in spatial transcriptomics, as it ensures that spatial relationships between gene expressions across tissue regions are better captured. SEPAL, in particular, demonstrates the highest PCC improvement with a 188\% increase, highlighting its ability to better align spatial gene expression patterns when training on \methodname{}-completed data.


\subsection{Ablation Experiments}

To better understand the effectiveness of our proposed method, we conduct ablation experiments using a random selection of six datasets from SpaRED. 
First, we evaluate the impact of removing context genes from our input data, meaning that we rely solely on the \specialgenes{} in their latent representation. Then, we explore working without the autoencoder component, directly training the diffusion branch on gene expression values. 

\begin{table}[h]
\centering
\caption{Average MSE in ablation experiments.}
\begin{tabular}{@{}cccc@{}}
\toprule
\multicolumn{1}{l}{\textbf{Latent space} \hspace{10 mm}} & 
\multicolumn{1}{l}{\textbf{Context genes} \hspace{10 mm}} & 
\multicolumn{1}{l}{\textbf{Average MSE}} \\ 
\midrule
\checkmark & \checkmark & \textbf{0.573 ± 0.136} \\
\checkmark & & 0.710 ± 0.214 \\
 & & 1.007 ± 0.200\\ 
\bottomrule
\end{tabular}
\label{tab:ablation}
\end{table}

Table \ref{tab:ablation} presents the results of the ablation experiments, demonstrating a 23\% decline in average performance when neighborhoods contain only the \specialgenes{} instead of using a latent space that includes context genes. Removing also the autoencoder component and working directly on gene expression values leads to a 62\% increase in completion error compared to removing only the context genes. 

\begin{table}[]
\centering
\caption{Average MSE with varying number of neighbors for \methodname{}. We present the number of
floating point operations (FLOPs) of each neighborhood configuration.}
\label{tab:num_neighs}
\begin{tabular}{cccl}
\hline
\multicolumn{1}{l}{Neighbors} & Average MSE            & \multicolumn{1}{l}{FLOPs (G)} \\ \hline
0                             & 1.891 ± 0.988          & 0.64  \\
6                             & \textbf{0.573 ± 0.136} & 3.22  \\
18                            & 0.588 ± 0.155          & 8.74 \\ \hline
\end{tabular}
\end{table}

As a final ablation experiment, we evaluate the complete \methodname{} model (including both the autoencoder and context genes) under three neighborhood size configurations. Our default setup uses six neighbors, equivalent to 1-hop in the hexagonal geometry of Visium. We also examine the effects of removing neighborhood conditioning altogether, as well as expanding the neighborhood to 18 neighbors (2-hops). The results in Table~\ref{tab:num_neighs} underscore the importance of neighboring spots for accurate predictions, considering that removing neighborhood information increases the MSE by 230\%. Although expanding to a 2-hop neighborhood yields similar performance to 1-hop, it increases the number of FLOPs by 171\%, resulting in a substantially higher computational cost without performance improvement.


These findings indicate that including known data from neighboring spots successfully guides the completion process and that latent representations provide biologically meaningful information for data completion. Similarly, context genes prove to be valuable for certain ST analyses, despite having a low spatial autocorrelation. While \methodname{} is computationally expensive by design, with a diffusion model backbone that requires 3.22 GFLOPs, restricting the neighborhood to 1-hop achieves strong performance without substantially increasing the computational cost.


%% file: sec/5_conclusions.tex
\section{Conclusions}
In this paper, we present \methodname{}, a reference-free ST completion model that effectively addresses the challenges of tackling data dropout. We show the suitability of generative models combined with AEs to predict missing values in ST and propose a novel approach to harness the contextual information available in low spatially autocorrelated genes. Our results prove that \methodname{} outperforms the previous state-of-the-art with an average MSE reduction of 18\% across the SpaRED database. Finally, we demonstrate that completing ST data with \methodname{} improves its reliability and unlocks the full potential of this sequencing technology. By improving the performance of gene expression prediction models when training them on \methodname{}-completed data, we highlight its broader impact on advancing histopathology image analysis and related biomedical research.

%% file: sec/6_acknowledgments.tex
\section{Acknowledgments}
Daniela Vega acknowledges the support of a UniAndes Google-DeepMind 2024 scholarship. This work was supported by Azure sponsorship credits granted by Microsoft’s AI for Good Research Lab.

%% file: main.bbl
\begin{thebibliography}{31}
\providecommand{\natexlab}[1]{#1}
\providecommand{\url}[1]{\texttt{#1}}
\expandafter\ifx\csname urlstyle\endcsname\relax
  \providecommand{\doi}[1]{doi: #1}\else
  \providecommand{\doi}{doi: \begingroup \urlstyle{rm}\Url}\fi

\bibitem[Abdelaal et~al.(2020)Abdelaal, Mourragui, Mahfouz, and Reinders]{spage}
Tamim Abdelaal, Soufiane Mourragui, Ahmed Mahfouz, and Marcel~JT Reinders.
\newblock Spage: spatial gene enhancement using scrna-seq.
\newblock \emph{Nucleic acids research}, 48\penalty0 (18):\penalty0 e107--e107, 2020.

\bibitem[Berahmand et~al.(2024)Berahmand, Daneshfar, Salehi, Li, and Xu]{berahmand2024autoencoders}
Kamal Berahmand, Fatemeh Daneshfar, Elaheh~Sadat Salehi, Yuefeng Li, and Yue Xu.
\newblock Autoencoders and their applications in machine learning: a survey.
\newblock \emph{Artificial Intelligence Review}, 57\penalty0 (2):\penalty0 28, 2024.

\bibitem[Biancalani et~al.(2021)Biancalani, Scalia, Buffoni, Avasthi, Lu, Sanger, Tokcan, Vanderburg, Segerstolpe, Zhang, et~al.]{tangram}
Tommaso Biancalani, Gabriele Scalia, Lorenzo Buffoni, Raghav Avasthi, Ziqing Lu, Aman Sanger, Neriman Tokcan, Charles~R Vanderburg, {\AA}sa Segerstolpe, Meng Zhang, et~al.
\newblock Deep learning and alignment of spatially resolved single-cell transcriptomes with tangram.
\newblock \emph{Nature methods}, 18\penalty0 (11):\penalty0 1352--1362, 2021.

\bibitem[Cheng et~al.(2023)Cheng, Jiang, Xu, Mentis, Wang, Zheng, Sahu, Liu, and Xu]{cheng2023spatially}
Mengnan Cheng, Yujia Jiang, Jiangshan Xu, Alexios-Fotios~A Mentis, Shuai Wang, Huiwen Zheng, Sunil~Kumar Sahu, Longqi Liu, and Xun Xu.
\newblock Spatially resolved transcriptomics: a comprehensive review of their technological advances, applications, and challenges.
\newblock \emph{Journal of Genetics and Genomics}, 50\penalty0 (9):\penalty0 625--640, 2023.

\bibitem[Defard et~al.(2024)Defard, Laporte, Ayan, Soulier, Curras-Alonso, Weber, Massip, Londo{\~n}o-Vallejo, Fouillade, Mueller, et~al.]{defard2024point}
Thomas Defard, Hugo Laporte, Mallick Ayan, Juliette Soulier, Sandra Curras-Alonso, Christian Weber, Florian Massip, Jos{\'e}-Arturo Londo{\~n}o-Vallejo, Charles Fouillade, Florian Mueller, et~al.
\newblock A point cloud segmentation framework for image-based spatial transcriptomics.
\newblock \emph{Communications Biology}, 7\penalty0 (1):\penalty0 823, 2024.

\bibitem[Haque et~al.(2017)Haque, Engel, Teichmann, and L{\"o}nnberg]{dropout}
Ashraful Haque, Jessica Engel, Sarah~A Teichmann, and Tapio L{\"o}nnberg.
\newblock A practical guide to single-cell rna-sequencing for biomedical research and clinical applications.
\newblock \emph{Genome medicine}, 9:\penalty0 1--12, 2017.

\bibitem[He et~al.(2020)He, Bergenstr{\aa}hle, Stenbeck, Abid, Andersson, Borg, Maaskola, Lundeberg, and Zou]{stnet}
Bryan He, Ludvig Bergenstr{\aa}hle, Linnea Stenbeck, Abubakar Abid, Alma Andersson, {\AA}ke Borg, Jonas Maaskola, Joakim Lundeberg, and James Zou.
\newblock Integrating spatial gene expression and breast tumour morphology via deep learning.
\newblock \emph{Nature biomedical engineering}, 4\penalty0 (8):\penalty0 827--834, 2020.

\bibitem[Ho et~al.(2020)Ho, Jain, and Abbeel]{ddpms}
Jonathan Ho, Ajay Jain, and Pieter Abbeel.
\newblock Denoising diffusion probabilistic models.
\newblock \emph{Advances in neural information processing systems}, 33:\penalty0 6840--6851, 2020.

\bibitem[Jin et~al.(2024)Jin, Zuo, Li, Liu, Pan, Fan, Fu, Yao, and Peng]{jin2024advances}
Yang Jin, Yuanli Zuo, Gang Li, Wenrong Liu, Yitong Pan, Ting Fan, Xin Fu, Xiaojun Yao, and Yong Peng.
\newblock Advances in spatial transcriptomics and its applications in cancer research.
\newblock \emph{Molecular Cancer}, 23\penalty0 (1):\penalty0 129, 2024.

\bibitem[Kingma and Ba(2014)]{kingma2014adam}
Diederik~P Kingma and Jimmy Ba.
\newblock Adam: A method for stochastic optimization.
\newblock \emph{arXiv preprint arXiv:1412.6980}, 2014.

\bibitem[Komura et~al.(2024)Komura, Ochi, and Ishikawa]{komura2024machine}
Daisuke Komura, Mieko Ochi, and Shumpei Ishikawa.
\newblock Machine learning methods for histopathological image analysis: Updates in 2024.
\newblock \emph{Computational and Structural Biotechnology Journal}, 2024.

\bibitem[Li et~al.(2024)Li, Li, Tao, and Wang]{stDiff}
Kongming Li, Jiahao Li, Yuhao Tao, and Fei Wang.
\newblock stdiff: a diffusion model for imputing spatial transcriptomics through single-cell transcriptomics.
\newblock \emph{Briefings in Bioinformatics}, 25\penalty0 (3):\penalty0 bbae171, 2024.

\bibitem[Li et~al.(2022)Li, Stanojevic, and Garmire]{li2022emerging}
Yijun Li, Stefan Stanojevic, and Lana~X Garmire.
\newblock Emerging artificial intelligence applications in spatial transcriptomics analysis.
\newblock \emph{Computational and Structural Biotechnology Journal}, 20:\penalty0 2895--2908, 2022.

\bibitem[Luo et~al.(2024)Luo, Hao, Wei, and Zhang]{scDiffusion}
Erpai Luo, Minsheng Hao, Lei Wei, and Xuegong Zhang.
\newblock scdiffusion: conditional generation of high-quality single-cell data using diffusion model.
\newblock \emph{Bioinformatics}, 40\penalty0 (9):\penalty0 btae518, 2024.

\bibitem[Marel(2024)]{marel2024navigating}
Ricfrid van~der Marel.
\newblock Navigating the complexity of data imputation in spatial transcriptomics: Strategies, challenges, and future directions.
\newblock 2024.

\bibitem[Marx(2021)]{marx2021method}
Vivien Marx.
\newblock Method of the year: spatially resolved transcriptomics.
\newblock \emph{Nature methods}, 18\penalty0 (1):\penalty0 9--14, 2021.

\bibitem[Mejia et~al.(2023)Mejia, C\'ardenas, Ruiz, Castillo, and Arbel\'aez]{sepal}
Gabriel Mejia, Paula C\'ardenas, Daniela Ruiz, Angela Castillo, and Pablo Arbel\'aez.
\newblock Sepal: Spatial gene expression prediction from local graphs.
\newblock In \emph{Proceedings of the IEEE/CVF International Conference on Computer Vision (ICCV) Workshops}, pages 2294--2303, 2023.

\bibitem[Mejia et~al.(2024)Mejia, Ruiz, C{\'a}rdenas, Manrique, Vega, and Arbel{\'a}ez]{spackle}
Gabriel Mejia, Daniela Ruiz, Paula C{\'a}rdenas, Leonardo Manrique, Daniela Vega, and Pablo Arbel{\'a}ez.
\newblock Enhancing gene expression prediction from histology images with spatial transcriptomics completion.
\newblock In \emph{International Conference on Medical Image Computing and Computer-Assisted Intervention}, pages 91--101. Springer, 2024.

\bibitem[Moran(1950)]{moran}
Patrick~AP Moran.
\newblock Notes on continuous stochastic phenomena.
\newblock \emph{Biometrika}, 37\penalty0 (1/2):\penalty0 17--23, 1950.

\bibitem[Pang et~al.(2021)Pang, Su, and Li]{histogene}
Minxing Pang, Kenong Su, and Mingyao Li.
\newblock Leveraging information in spatial transcriptomics to predict super-resolution gene expression from histology images in tumors.
\newblock \emph{BioRxiv}, pages 2021--11, 2021.

\bibitem[Peebles and Xie(2023)]{dit}
William Peebles and Saining Xie.
\newblock Scalable diffusion models with transformers.
\newblock In \emph{Proceedings of the IEEE/CVF international conference on computer vision}, pages 4195--4205, 2023.

\bibitem[Pham et~al.(2020)Pham, Tan, Xu, Grice, Lam, Raghubar, Vukovic, Ruitenberg, and Nguyen]{stLearn}
Duy Pham, Xiao Tan, Jun Xu, Laura~F Grice, Pui~Yeng Lam, Arti Raghubar, Jana Vukovic, Marc~J Ruitenberg, and Quan Nguyen.
\newblock stlearn: integrating spatial location, tissue morphology and gene expression to find cell types, cell-cell interactions and spatial trajectories within undissociated tissues.
\newblock \emph{biorxiv}, pages 2020--05, 2020.

\bibitem[Rombach et~al.(2022)Rombach, Blattmann, Lorenz, Esser, and Ommer]{LDM}
Robin Rombach, Andreas Blattmann, Dominik Lorenz, Patrick Esser, and Bj{\"o}rn Ommer.
\newblock High-resolution image synthesis with latent diffusion models.
\newblock In \emph{Proceedings of the IEEE/CVF conference on computer vision and pattern recognition}, pages 10684--10695, 2022.

\bibitem[Shengquan et~al.(2021)Shengquan, Boheng, Xiaoyang, Xuegong, and Rui]{stPlus}
Chen Shengquan, Zhang Boheng, Chen Xiaoyang, Zhang Xuegong, and Jiang Rui.
\newblock stplus: a reference-based method for the accurate enhancement of spatial transcriptomics.
\newblock \emph{Bioinformatics}, 37\penalty0 (Supplement\_1):\penalty0 i299--i307, 2021.

\bibitem[Soleymani et~al.(2024)Soleymani, Paquet, Viktor, and Michalowski]{soleymani2024structure}
Farzan Soleymani, Eric Paquet, Herna~Lydia Viktor, and Wojtek Michalowski.
\newblock Structure-based protein and small molecule generation using egnn and diffusion models: A comprehensive review.
\newblock \emph{Computational and Structural Biotechnology Journal}, 2024.

\bibitem[Wang et~al.(2023)Wang, Liu, Zhao, Lee, Buzdin, Mu, Zhao, Chen, and Li]{wang2023spatial}
Ye Wang, Bin Liu, Gexin Zhao, YooJin Lee, Anton Buzdin, Xiaofeng Mu, Joseph Zhao, Hong Chen, and Xinmin Li.
\newblock Spatial transcriptomics: Technologies, applications and experimental considerations.
\newblock \emph{Genomics}, 115\penalty0 (5):\penalty0 110671, 2023.

\bibitem[Xie et~al.(2023)Xie, Pang, Chung, Perciani, MacParland, Wang, and Bader]{bleep}
Ronald Xie, Kuan Pang, Sai Chung, Catia Perciani, Sonya MacParland, Bo Wang, and Gary Bader.
\newblock Spatially resolved gene expression prediction from histology images via bi-modal contrastive learning.
\newblock \emph{Advances in Neural Information Processing Systems}, 36:\penalty0 70626--70637, 2023.

\bibitem[Yan et~al.(2024)Yan, Zhu, Chen, Yang, Cui, Zou, and Zhang]{yan2024integration}
Chaorui Yan, Yanxu Zhu, Miao Chen, Kainan Yang, Feifei Cui, Quan Zou, and Zilong Zhang.
\newblock Integration tools for scrna-seq data and spatial transcriptomics sequencing data.
\newblock \emph{Briefings in Functional Genomics}, 23\penalty0 (4):\penalty0 295--302, 2024.

\bibitem[Yang et~al.(2023)Yang, Hossain, Stone, and Rahman]{egn}
Yan Yang, Md~Zakir Hossain, Eric~A Stone, and Shafin Rahman.
\newblock Exemplar guided deep neural network for spatial transcriptomics analysis of gene expression prediction.
\newblock In \emph{Proceedings of the IEEE/CVF Winter Conference on Applications of Computer Vision}, pages 5039--5048, 2023.

\bibitem[Yang et~al.(2024)Yang, Hossain, Stone, and Rahman]{yang2024spatial}
Yan Yang, Md~Zakir Hossain, Eric Stone, and Shafin Rahman.
\newblock Spatial transcriptomics analysis of gene expression prediction using exemplar guided graph neural network.
\newblock \emph{Pattern Recognition}, 145:\penalty0 109966, 2024.

\bibitem[Zhu et~al.(2025)Zhu, Zhu, Tao, and Qiu]{stem}
Sichen Zhu, Yuchen Zhu, Molei Tao, and Peng Qiu.
\newblock Diffusion generative modeling for spatially resolved gene expression inference from histology images.
\newblock \emph{arXiv preprint arXiv:2501.15598}, 2025.

\end{thebibliography}
